\documentclass[11pt,a4paper]{article}
\usepackage[hyperref]{acl2018}
\usepackage{times}
\usepackage{latexsym}

\usepackage{url}

\aclfinalcopy 

\title{On the Compression of Natural Language Models}

\author{%
  Saeed Damadi \\
  Department of Computer Science and Electrical Engineering Department\\
  University of Maryland, Baltimore County\\
  \texttt{sdamadi1@umbc.edu} \\
}

\date{}

\begin{document}
\maketitle
\begin{abstract}
Deep neural networks are effective feature extractors but they are prohibitively large for deployment scenarios.
Due to the huge number of parameters, interpretability of parameters in different layers is not straight-forward.
This is why neural networks are sometimes considered black boxes. 
Although simpler models are easier to explain, finding them is not easy. If found, a sparse network that can fit to a data from scratch would help to interpret parameters of a neural network.
To this end, \cite{frankle2018lottery} showed that typical dense neural networks contain a small sparse sub-network that can be trained to a reach similar test accuracy in an equal number of steps.
The goal of this work is to assess whether such a trainable subnetwork exists for natural language models (NLM)s. To achieve this goal we will review state-of-the-art compression techniques such as  quantization, knowledge distillation, and pruning.
\end{abstract}

\section{Introduction}

Although deep neural networks (DNNs) are over-parameterized \cite{denil2013predicting}, they have been able to solve many difficult problems. For example, pre-trained feature extractors, such as BERT \cite{devlin2018bert} for natural language processing and residual networks for computer vision tasks, have become effective methods for improving deep learning models without requiring more labeled data. However, they are sizable and cannot be deployed on devices with small memories like smart phones. This has had researchers to compress DNNs. 
To compress DNNs there are three main ways: 1) quantization, 2) knowledge distillation, and 3) pruning where the last one is the main focus of this manuscript. The less parameters network has, the less computational cost would be incurred. This cost may be associated with training or testing a DNN. However, most of compression techniques try to reduce computational cost associated with running queries (testing). This is justified by stating that ``training is done once, and testing is done infinitely many times''. In addition to reducing the number of parameters, one may reduce the accuracy of the parameters in order not to occupy the memory. This is the most straight forward way for reducing computational cost of testing which is called quantization and will be explained briefly as follows.

\subsection{Quantization}
Quantization techniques try to store model parameters from 32- or 16-bit floating number to 8-bit or even lower. This is a very simple method for reducing the computational cost of queries. Although, from implementation perspective is very effective, reducing the bit representation of parameters may cause significant accuracy loss. To avoid that, quantization-aware training has been developed to maintain similar accuracy to the original model \cite{shen2020q,zafrir2019q8bert}.
Quantization is applied on a fixed DNN after training. On the other hand, there are techniques looking for a different smaller structures that can learn the same task. One of these method is knowledge distillation which is explained as follows.

\subsection{Knowledge distillation}
In essence, knowledge distillation tries to use a large trained network to train a smaller network. In this method, a small network is forced to have an output that is similar to a large trained network. The large trained network is called \textit{teacher} and the small network is called \textit{student}.
Therefore, a compressed and shallow student network under the guidance of a complicated larger teacher network is trained. The trained compressed student network can be directly deployed in real-life applications.
Existing methods can be generally divided into two categories: 1) task-specific, and 2) task-agnostic.
Task-specific methods require that the \textit{teacher} be trained for each downstream task. Distilled bidirectional long short-term memory network (Distilled BiLSTM) \cite{tang2019distilling}, Patient Knowledge Distillation for a BERT model (BERT-PKD) \cite{sun2019patient}, and Stacked Internal Distillation (SID) \cite{aguilar2020knowledge} are all considered as task-specific methods. On the other hand, task-agnostic method uses one \textit{teacher} for several downstream tasks. In effect, one \textit{teacher} can train multiple \textit{students}. Methods such as DistilBERT \cite{sanh2019distilbert},
TinyBERT \cite{jiao2019tinybert} and MobileBERT \cite{sun2020mobilebert}  are task-specific methods.
Note that process of task-agnostic BERT distillation is computationally expensive \cite{mccarley2019structured} because the corpus used in the
distillation is sizable and for each training step a forward process of \textit{teacher} model and a forward-backward process
of \textit{student} model should be performed.

\subsubsection{The larger the \textit{teacher}, the better}
The more information the \textit{teacher} can provide with the the \textit{student}, the better the performance of the \textit{student} model would be. To this end, TinyBERT, MobileBERT, and SID all try to improve BERT-PKD by distilling more internal representations to the \textit{student}, such as embedding layers and attention weights. 
TinyBERT \cite{jiao2019tinybert} and MobileBERT \cite{sun2020mobilebert} are small \textit{student} models distilled from larger pre-trained transformers and can achieve good GLUE \cite{wang2018glue}
scores. However, these models require spending substantial compute
to pre-train the larger \textit{teacher} model. 
To address this issue ELECTRA-Small \cite{clark2020electra} focus more on pre-training speed rather than inference speed.
\subsection{Pruning}
The main focus of this paper is pruning as a method for compressing neural networks. Pruning refers to identifying and removing redundant or less important parameters.
There are two approaches for pruning NLMs: 1) structured, and 2) unstructured. The main difference between the two is that the former drops network block(s), while the latter zeros out individual parameters across the entire network.

\subsubsection{Structured Pruning}
In structured pruning, architecture blocks (layers) are pruned.
Structured pruning first utilized for convolutional neural networks but it has recently been applied to NLMs. Structure pruning can be done in two directions, depth (layers) or width (headers) of the network.
In the depth direction, pruning transformer layers is proposed in LayerDrop \cite{fan2019reducing}
via structured dropout.
\cite{sajjad2020poor} prunes the network in the depth direction as well.
In the width direction \cite{michel2019sixteen,voita2019analyzing,mccarley2019structured} retain the performance after pruning a large percentage of attention heads in a structured manner.
\subsubsection{Unstructured Pruning} 
In unstructured pruning, parameters with the least importance are removed from the network. The importance of the
weights can be judged by their absolute values, the gradients, or by some custom-designed measurement.  Since unstructured pruning considers each parameter individually, the set
of pruned ones can be arbitrary and irregular,
which in turn might decrease the model size, but with negligible improvement in runtime memory
or inference speedup \cite{sanh2020movement}, unless executed on specialized hardware or with specialized processing libraries \cite{han2016eie,chen2018escoin}.
Unstructured pruning could be
effective for BERT, given the massive amount of fully-connected layers. 

\subsection{State-of-the-art-pruning methods}
In this subsection we will go over state-of-the-art-pruning methods. As the goal is to explore as many as methods, we briefly introduce each method and do not dive into them. 

\subsubsection{Pruning of question answering models}
To create small task-specific language representations \cite{mccarley2019structured} tries to compress BERT- and RoBERTa \cite{liu2019roberta} based question answering systems by combining task-specific knowledge distillation and task-specific structured pruning. The intuition is based on two observations: 1) a general-purpose language representation requires expensive pretraining distillation so the number of tasks should be limited, and 2) BERT models are robust to attention heads pruning.
\subsection{LayerDrop: Structured dropout}
Dropping an entire layer which is considered as structured pruning is studied in \cite{fan2019reducing}. They show that it is possible to extract shallower subnetworks from large networks without having to finetune. The performance of these subnetworks is close to the dense network. 
DropLayer \cite{fan2019reducing} uses group regularization
to enable structured pruning at inference time.
\subsubsection{Head pruning}
The original BERT model
has 16 attention heads and \cite{michel2019sixteen} questions the number of heads in the BERT model. \cite{michel2019sixteen} shows that many of
BERT’s attention heads can be pruned, while high accuracy during test time is possible with only 1–2 attention heads per encoder
unit. This fact shows that there is a  high  redundancy in the learned parameters of BERT model. Pruning these many attention heads improve memory efficiency in transformers.

\subsubsection{Factorized low-rank Pruning}
By combining pruning and matrix factorization for model compression
\cite{wang2019structured} proposes a method to structurally prune BERT models using low-rank factorization and augmented Lagrangian $L_0$ norm regularization. The method is called Factorized Low-rank Pruning (FLOP).
Given $W$ as a weight matrix, structured pruning can be achieved by replacing the computation of $Wx$ by $WGx$ where diagonal sparsity inducing matrix $G$ is learned using $L_0$ regularization over $WG$ along with the supervised loss. Next, $W$ is decomposed to two smaller matrix $P$ and $Q$ such that $W=PGQ$. In other words \cite{wang2019structured}  reparameterizes weight matrices using low-rank factorization and removes rank-1 components during training. One limitation of this method is that this structured pruning method tends to produce lower performance than its unstructured counterpart.
\subsubsection{Reweighted proximal pruning}
Observing the fact that larger weights are penalized more heavily than smaller ones when $L_1$ regularization is utilized, \cite{guo2019reweighted} proposes  reweighted $L_1$ minimization. This observation is the intrinsic artifact of $L_1$ minimization. Once it happens it defeats the purpose of  weight pruning which is “removing the unimportant connections”. \cite{guo2019reweighted} observes that direct optimization of a regularization penalty term causes divergence from the original loss function and has negative effect on the effectiveness of gradient-based update. To avoid aforementioned problems they solve the following optimization problem:
$$
\min_{w} f(w)+\sum_i \alpha_i w_i
$$
where $f(w)$ is the original loss function, $w$ is a parameter (weight) and $0<\alpha_i$'s are inversely proportional to magnitude of corresponding weights $w_i$. This optimization is solved using a reweighted proximal
pruning (RPP) method (which depends on proximal operators). RPP decouples the goal of high sparsity from minimizing
loss, and hence leads to improved accuracy even with high levels of pruning for BERT models.
\subsubsection{Compressing pretrained BERT}
Using pretrained BERT as feature extractor is widespread for natural language processing tasks. Because their learned parameters can be transfer to other tasks. 
\cite{gordon2020compressing} explores the effects of unstructured pruning on transfer learning. \cite{gordon2020compressing} observes that low level of pruning (30-40\%) let learned parameters to be transformed to downstream tasks without any increase in the loss. As opposed to low level pruning, medium and high levels of pruning prevent useful pretraining information be transferred to downstream tasks. For example, Multi-Genre Natural Language Inference (MNLI) is the task that is most sensitive to pruning under their method.
The main observation of  \cite{gordon2020compressing} is that BERT
can be pruned during pretraining rather than separately for each task without affecting
performance. 
\subsubsection{Movement pruning}
\cite{sanh2020movement} argues that for transfer learning, what matters is not the magnitude of a
parameter. What matters is that whether that parameter is important for the downstream task. They introduce movement pruning
which is a first-order method. This method prunes parameters shrinking during fine-tuning regardless of their magnitude. Movement pruning is able to achieve significantly higher performance than magnitude- or $L_0$ based pruning for very high levels of sparsity (e.g., 97\% sparse). It also can be combined with distillation to improve the performance further.
\subsubsection{Know what you don’t
need}
\cite{zhang2021know} shows that many attention heads can be
removed without a significant impact on the performance.
To remove those attention heads a single-shot meta-pruner is introduced. This single-shot meta-pruner is a small convolutional neural network that is trained to select heads that contribute to maintaining the
attention distribution.
\subsection{The Lottery Ticket Hypothesis}
The Lottery Ticket Hypothesis (LTH) states that typical dense neural networks contain a small sparse sub-network that can be trained to reach similar test accuracy in an equal number of steps \cite{frankle2018lottery}. In view of that, follow-up works reveal that sparsity patterns might emerge at the initialization
\cite{lee2018snip}, the early stage of training 
\cite{you2019drawing} and \cite{chen2020earlybert}, or in dynamic forms throughout training \cite{evci2020rigging} by
updating model parameters and architecture typologies simultaneously.
Some of the recent findings are that the lottery
ticket hypothesis holds
for BERT models, i.e., largest weights of the original network do form subnetworks that can be retrained alone to reach the performance close to that of the full model \cite{prasanna2020bert,chen2020lottery}.
\subsubsection{LTH for BERT}
\cite{chen2020lottery} applies the LTH to identify matching subnetworks in pretrained BERT  models to enforce  sparsity  in models trained for different  downstream tasks.
They show that assuming pretrained parameters as the initialization for BERT models, one can find sparse subnetworks that can be trained for downstream NLP tasks. They also show that using unstructured magnitude pruning, one can find matching subnetworks at between 40\% and
90\% sparsity in BERT models on standard GLUE and SQuAD downstream tasks. While this approach retains accuracy and can lead to  sparser  models, it does  not lead to improvements in training speed  without dedicated hard-ware or libraries \cite{han2016eie,chen2018escoin}. On the other hand,  in structured pruning, the best subnetworks of BERT’s heads do not quite reach the full model performance.
\cite{chen2020lottery} shows for a range of downstream tasks, matching subnetworks at 40\% to 90\% sparsity exist and they are found at pretrained phase (initialization). This is dissimilar to the prior NLP research where subnetworks emerge only after some amount of training. Subnetworks that are found on the masked language modeling task transfer universally; those found on other tasks transfer in a limited fashion if at all.

\subsubsection{EarlyBERT}
\cite{chen2020earlybert} introduces EarlyBERT which extends the work done on finding lottery-tickets in CNNs \cite{you2019drawing} to speedup both pre-training
and fine-tuning for BERT models. \cite{you2019drawing} realized that sparsity patterns might emerge at the initialization. Experimental evaluation of EarlyBERT shows some degradation in accuracy for fine-tuning.

\subsection{Conclusion}
\cite{prasanna2020bert} and \cite{chen2020lottery}
explore BERT models from the perspective of the lottery ticket hypothesis \cite{frankle2018lottery},
looking specifically at the ‘‘winning’’ subnetworks in pre-trained BERT. They find that such subnetworks do exist, and that transferability between subnetworks for different tasks varies. The two papers provide complementary results for magnitude pruning and use a task specific classifier which is randomly initialized.
\subsubsection{Issues}
The issue with pruning is that existing works on pruning of BERT yields inferior results than its small-dense counterparts such as TinyBERT \cite{jiao2019tinybert}.
While unstructured pruning may need dedicated hard-ware or libraries \cite{han2016eie,chen2018escoin}, one may prefer them over structured pruning because they maintain accuracy and can lead to  sparser models.
\subsubsection{An open question}
There is an open question as follows. Is there a sparse task-specific classifier that can be combined with a pruned pretrained BERT model to get the accuracy on a par with a dense trained network? If such a task-specific classifier exists, can we use it for other tasks?
\subsubsection{Research directions}
In computer vision, \cite{kusupati2020soft} sparsifies the network with reparameterization technique that uses one extra single parameters for each layer. This method can be applied to a pretrained BERT to see how well it performs. 

\newpage
\bibliography{acl2018}
\bibliographystyle{acl_natbib}

\end{document}